\documentstyle[psfig]{llncs}

\def\ie{i.e.}
\def\eg{e.g.}

\def\la{$\langle$}
\def\ra{$\rangle$}

\def\cl{{\bf IN}}
\def\Sc{{\bf SC}}

\def\gcl{{\tt Greedy-IN}}
\newstytheorem{myexample}{}{\footnotesize}[example]{}
\begin{document}
\title{Classification in Feature-based Default Inheritance
  Hierarchies\thanks{This work was funded by NSF grant IRI-9013160,
    ONR/DARPA research grant no. N00014-92-J-1512, and a grant from
    the Land Baden-W\"{u}rttemberg, Germany.} }

\author{Marc Light}

\institute{
University of Rochester, Computer Science Dept. \\
Rochester, NY 14627 USA, {\tt light@cs.rochester.edu}
}

\maketitle

\begin{abstract}
  Increasingly, inheritance hierarchies are being used to reduce
  redundancy in natural language processing lexicons.  Systems that
  utilize inheritance hierarchies need to be able to insert words
  under the optimal set of classes in these hierarchies.  In this
  paper, we formalize this problem for feature-based default
  inheritance hierarchies.  Since the problem turns out to be
  NP-complete, we present an {\em approximation\/} algorithm for
  it.  We show that this algorithm is efficient and that it performs
  well with respect to a number of standard problems for default
  inheritance.  A prototype implementation has been tested on lexical
  hierarchies and it has produced encouraging results.  The work
  presented here is also relevant to other types of default
  hierarchies.
\end{abstract}
{\def\abstractname{Zusammenfassung.}
\begin{abstract}
  In zunehmendem Masse werden
  Erbschaftshierarchien zur kompakten Beschreibung von
  Worteigenschaften in Sprachverarbeitungslexica verwendet.  Systeme,
  die Erbschaftshierarchien ben\"{u}tzen, m\"{u}ssen Worte in die
  optimale Klasse (oder Menge von Klassen) der Hierarchien
  einf\"{u}gen k\"{o}nnen.  Dieser Beitrag formalisiert das Problem
  f\"{u}r merkmalsbasierte {\em default\/}-Hierarchien.  Da das
  Problem NP-vollst\"{a}ndig ist, wird hier ein
  Approximationsalgorithmus vorgeschlagen.  Es wird gezeigt, da\ss\ 
  der Algorithmus effizient ist und da\ss\ er in Hinsicht auf einige
  Standardprobleme der {\em default-\/}Hierarchien gut funktioniert.
  Eine Prototypimplementierung zeigt gute Resultate an einigen
  Hierarchien, die f\"{u}r Lexica der Computerlinguistik geschrieben
  wurden.  Diese Forschung kann auch f\"{u}r andere Arten von {\em
    default\/}-Hierarchien angewendet werden.
\end{abstract}}

\section{Introduction}

Recent computational linguistics research of natural language lexicons
has gone beyond simply listing idiosyncratic information about words.
Many researchers are adding structure to their lexicons to replace the
old practice of simply listing, in an entry for a word, all its
properties.  This structure often takes the form of feature-based
default inheritance hierarchies (\eg\ \cite{briscoe93,evens90}).
After such hierarchies have been designed, the task of placing word
entries under the appropriate classes begins.  The problem we address
in this paper is deciding where to place a new object in an
inheritance hierarchy---we will call this `the insertion problem.'
This problem has received little attention in the literature; to our
knowledge this paper is the first to formalize the problem and to
provide a general algorithm for it.
\begin{figure}[htbp]
% \centerline{\psfig{figure=/u/light/inserter/aaai/flick.ps,height=2.5in}}
  \centerline{\psfig{figure=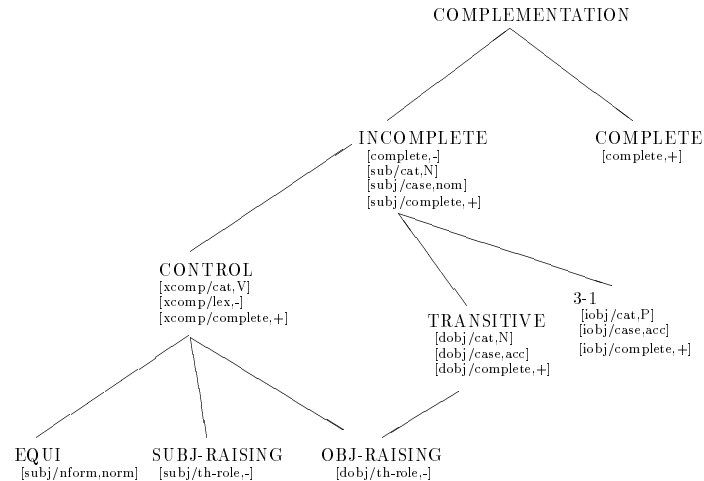,height=2.6in}}%2.5
  \caption{Verb feature hierarchy (adapted from
    \protect\cite{flickinger87})} 
  \label{flick}
\end{figure}

An example of the insertion problem is the task of placing a verb into
the hierarchy presented in Fig.\ref{flick}.  The upper case symbols
(\eg , CONTROL) are class names and the pairs\footnote{Both elements
  of these pairs are atomic symbols.  Here and throughout this paper,
  the `/' has only mnemonic relevance.} written directly below these
names represent features that can be inherited from them.  For
example, all the features of INCOMPLETE are inherited by EQUI---unless
overridden by features in CONTROL or in EQUI.  Consider the form of
{\em give\/} exemplified in {\em she gave him the ball\/} and assume
that it has the following features.
%\enumsentence{\label{givefeats}
%\footnotesize{
\begin{myexample}\label{givefeats}
\begin{tabbing}
aaa\= \kill
\>
[subj/cat,N] 
[subj/case,nom] 
[subj/complete,+] 
[dobj/cat,N] \\\null
\>
[dobj/case,acc] 
[dobj/complete,+] 
[iobj/cat,N] 
[iobj/case,acc] 
[iobj/complete,+] 
\end{tabbing} 
\end{myexample}
A solution to this instance of the insertion problem would be to place
{\em give\/} under the TRANSITIVE and 3-1 classes.  By doing so, all
of the features listed above could be inherited except [iobj/cat,N].
This feature would have to be listed directly in the entry for {\em
  give\/} so that the incorrect default inheritance of [iobj/cat,P]
from the 3-1 class would be blocked.  Thus, only three pieces of
information need to be listed in the entry for {\em give\/}: it
inherits from TRANSITIVE, it inherits from 3-1, and it has the feature
[iobj/cat,N].  Notice that if we had inserted the entry under any
other set of classes, more information would have had to be listed in
the entry.  A central characteristic of a good insertion is that it
minimizes the amount of information that needs to be listed in the
entry.

In this paper, we will focus on feature-based default inheritance
hierarchies: feature-based in that attribute/value pairs are used to
encode characteristics of a class and default in that all inheritance
relationships are defeasible.  We will also assume that multiple
inheritance is allowed: classes can inherit from more that one
superclass.  However, the results described in this paper are relevant
to {\bf any} system that utilizes defaults and for which inconsistency
between two elements that are inherited can be discovered efficiently.

The rest of this paper is structured as follows.  First, we give
an informal characterization of the insertion problem and argue that
this characterization captures the relevant aspects of the problem.
Second, we formalize this characterization. The problem, so defined,
is NP-complete \cite{light93d} and, thus, it is unlikely that a
computationally tractable algorithm for its solution will be found.
Next, we describe an approximation algorithm for the problem and
through complexity analysis and discussion of preliminary
experimentation, we argue that this algorithm produces reasonable
results in an acceptable amount of time and space.  The algorithm has
been implemented and has produced good results in experiments
involving inserting words into lexical hierarchies.

\section{\label{informal}Informal Characterization of the Problem}

As mentioned above, a good insertion places an object under classes
from which it can inherit most of its features.  In addition, it
should use as few classes as possible to achieve this goal.  These two
requirements are aspects of a basic principle of classification
schemes: reduce redundancy.  A corollary of this principle is that a
good insertion should minimize the amount of information stored in the
entry for the object being inserted and thereby maximize the use of
the information contained in the inheritance relationships.  Three
types of information are stored in an entry: its superclass(es), the
features that are not inherited from any superclass, and the features
that are needed to block an incorrect inheritance.  Thus, an optimal
insertion has the smallest possible value for the following sum (which
we will refer to as the cost of the insertion or solution).
%\enumsentence{\label{informsum} \footnotesize{
\begin{myexample}\label{informsum}
\begin{tabbing}
aaa\= \kill
\>
number of superclasses + number of object features not in these
superclasses + \\\null 
\>
the number of object features that must be listed to block incorrect
inheritance
\end{tabbing}
\end{myexample}

At first glance, one might think that one easy way to find an optimal
solution is to start at the roots of the hierarchies that make up a
structured lexicon and simply walk down these hierarchies, pruning off
the branches below a class that contains a feature that the object does
not have (since all classes below this class will also have this
feature).  This approach will work for {\em strict\/} inheritance
hierarchies.  Strict hierarchies do not allow default inheritance---all
the subclasses below a given class have to inherit all its features.
The algorithm outlined above uses this characteristic to cut down the
search space for superclasses.  However, default hierarchies do not
have this characteristic: if a class A inherits from a class B, the
set of features associated with A might not be a superset of the set
associated with B.  Thus, an insertion algorithm for default
hierarchies cannot ignore A as a possible superclass for an object
simply because it has decided B is unsuitable; A might be an exception
to B in just the right ways.
Because of these considerations, for the purposes of insertion, we make
the following claim: each class in a default hierarchy should be
viewed as the set of features that can be inherited from this class.
This claim is central to the approach to insertion taken in this
paper.  It amounts to `compiling out' the inheritance relationships so
that a hierarchy becomes a set of sets of features.  Compiling out
the inheritance relationships is a process of pushing features down
the inheritance links to the classes below so that all the features of
a class are {\em explicitly\/} listed in the data structure for the
class.  For example, compiling out the hierarchy in Fig.~\ref{flick}
would produce the set of sets listed in (\ref{compN}). The first set
corresponds to the COMPLEMENTATION class, the second to INCOMPLETE,
the third to COMPLETE, and the fourth to TRANSITIVE.
%\enumsentence{\label{compN}
%\footnotesize{
\begin{myexample}\label{compN}
\begin{tabbing}
aaa\= a\= a\= \kill
\>\{\{\} \\\null
\>\> \{[complete,-] [subj/cat,N] [subj/case,nom] [subj/complete,+]\} \\\null
\>\> \{[complete,+]\} \\\null
\>\> \{[complete,-] [subj/cat,N] [subj/case,nom] [subj/complete,+] \\\null
\>\> \> [dobj/cat,N] [dobj/case,acc] [dobj/complete,+]\} 
.... \}
\end{tabbing}
\end{myexample}

By compiling out a hierarchy, one loses inheritance relationships
between classes.  However, since a good insertion minimizes the space
needed to store the properties of the object being inserted, with
respect to insertion, a class is simply a chance to save space by
storing one class name instead of a number of features.  Thus, what
features can be inherited from a class is the only characteristic of a
class relevant to insertion.  As a result, when dealing with the
insertion problem in default hierarchies, one should think of the
hierarchy as a set of sets, each of which corresponds to a possible
superclass of the object being inserted.  The insertion problem, then,
amounts to picking an optimal subset of this set of sets; the
definition of optimal remains the same.

\section{\label{formal}Formal Definition}

In this section, we will formalize the intuitive characterization of
the previous section.  The first task is to give a formal
characterization of the features we have been discussing informally in
the previous section.  A feature is a pair of atomic symbols (\eg\ 
[a,v]).  The first element is taken from a set of attributes
(ATTRIBUTES) and the second from a set of values (VALUES); these sets
may be infinite.  The set VALUES includes a symbol {\tt ?} which
intuitively specifies that the corresponding attribute is undefined or
unknown for the object with the feature, more on this later.  Two
features clash (\ie , are inconsistent) if their attributes are the
same but their values are different; this definition also holds for
the value {\tt ?}.  Note that testing if two features clash can be
done in constant time.  We will use {\em clash\/} to denote a binary
function on sets of features that produces the subset of features from
the first set that are in conflict with a feature of the second (see
below).  
\pagebreak
%\enumsentence{\label{clash} \footnotesize{
\begin{myexample}\label{clash}
\begin{tabbing}
aaa\= \kill
\> A = \{[a1,v1][a2,v2][a3,v3][a4,v4]\} \\\null
\> B = \{[a1,v5][a2,v2][a3,v20][a7,v7][a9,v12]\} \\\null
\> {\em clash}(A,B) = \{[a1,v1][a3,v3]\} 
\end{tabbing} 
\end{myexample}
We assume that the sets of features are internally consistent, \ie ,
each set contains at most one feature per attribute.  $clash({\cal
  C},{\cal D})$ can be computed for the finite sets $\cal C$ and $\cal
D$ in $O(|{\cal C}||{\cal D}|)$ time by simply testing every pair
formed from an element from $\cal C$ and an element of $\cal D$ for a
clash.  

Although we will work with this specific feature system, the algorithms
and proofs that we present here, generalize to any system for
representing {\em what\/} is inherited with the following property:
whether two inheritable elements clash can be computed in time
polynomial in the size of the elements.  Features systems used in
natural language processing often have this property (\ie\
polynomial-time unification algorithm).

\begin{figure}[htbp]
% \centerline{\psfig{figure=/u/light/inserter/aaai/advcau.ps,height=1.4in}}
  \centerline{\psfig{figure=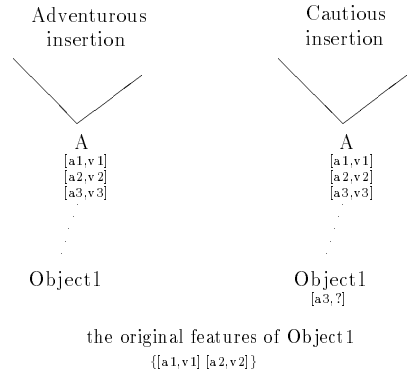,height=2in}} %1.8
  \caption{Adventurous vs. cautious insertion}
  \label{advcau}
\end{figure}

The next task is to further specify the hierarchies to be used.  As
mentioned above, we are concerned with feature-based default multiple
inheritance hierarchies. In addition, we assume that the hierarchies
are unambiguous: for any given attribute, a node only inherits one
value for it.  We will discuss the issue of ambiguity with respect to
insertion below.  We also assume, for expository reasons, that the
feature sets inheritable from the classes are unique: that a given set
of features cannot be inherited from more than one class in a
hierarchy.  Further, we are concerned with cautious insertion: only the
known features of the object being inserted should be inherited.  An
adventurous insertion would allow an object class to inherit extra
features from its superclasses.  In order to deal with the problem of
inheriting extra features from superclasses, we will use the value {\tt
  ?} in features to block unwanted inheritance.  we will call such
features `{\tt ?}-features.'  Only object classes are allowed to
contain {\tt ?}-features.  Fig.\ref{advcau} illustrates both forms
of insertion and the use of the {\tt ?} value.  In adventurous
insertion, Object1 would have the feature [a3,v3] whereas with
cautious insertion, its inheritance would be blocked by [a3,{\tt ?}].

In the previous section, we defined the insertion problem so that
solutions could be incomplete: not all the features of the object being
inserted need be inherited.  Of course, a solution pays a price
for forcing a feature to be listed in the object class.  The second
term of the sum in (\ref{informsum}) specifies the price paid.
Because of the possible existence of clashes (the third term), an
optimal solution may be incomplete.  Such incomplete insertions
complicate the reduction needed to prove NP-completeness and the
statement of the insertion algorithm.  To circumvent this difficulty,
we assume that all hierarchies contain a singleton class for each
non-{\tt ?}-feature of the object being inserted. (We will leave these
sets out of the figures of this paper.)  A singleton class is a class
from which exactly one feature can be inherited and which has no
superclasses and no subclasses other than object classes.  Listing a
singleton class as a superclass in an object class is equivalent to
simply listing the feature of the singleton class.  This assumption
allows us to require complete insertions while still being able to
represent the notion of incomplete insertion.

An instance of the insertion problem (\cl ) is a pair made up of a set
$\cal F$ and a set $\cal N$ as defined below.  Note that only a finite
number of features is needed to specify $\cal F$ and that there is a
one-to-one mapping between the sets in $\cal N$ and the classes of the
hierarchy.

\begin{description}
\footnotesize{
\item[$\cal F$] contains the features of the object being inserted.
  $\cal F$ is complete in the sense that for all attributes in
  ATTRIBUTES there exists a feature in $\cal F$ that contains a value
  for that attribute.  However, $\cal F$ only has a finite
  number of features that contain values other than {\tt ?}.  
\item[$\cal N$] is a set of sets of features.  $\cal N$ is finite
  and each of its elements is finite.  
}
\end{description}

If we return to our discussion of the hierarchy in Fig.\ref{flick},
we see that the set of features of {\em give\/}, listed in
(\ref{givefeats}), plus the appropriate {\tt ?}-features, is
$\cal{F}$ for this hierarchy.  The set of sets of features in
(\ref{compN}) that resulted from compiling out the hierarchy, plus
singleton sets for the features of $\cal{F}$, is $\cal{N}$.

A solution is a set ${\cal P} \subseteq {\cal N}$.  $\cal P$
represents the superclasses for the object whose features are $\cal
F$.  Every non-{\tt ?}-feature in $\cal F$ must be an element of
either a set in $\cal P$ or $clash({\cal F},\bigcup {\cal
  P})$.\footnote{$\bigcup \Phi$ where $\Phi$ is a set of sets
  $\varphi_{1},\varphi_{2},...,\varphi_{n}$ is equal to
  $\varphi_{1}\cup\varphi_{}\cup ... \cup\varphi_{n}$.} The features
in this latter set have to be listed explicitly for the object in
order to override incorrect inheritance.  An optimal solution
minimizes the number of superclasses together with the number of
clashes between these superclasses and $\cal F$: $|{\cal P}| +
|clash({\cal F},\bigcup {\cal P})|$.
Finally, the insertion problem, \cl , is the problem of
finding optimal solutions for instances as defined above.

\section{An Approximation Algorithm for \cl }

As mentioned above, \cl\ is NP-complete.  Thus, the best one should
hope for is a computationally tractable algorithm that produces {\em
  near\/} optimal solutions.  Part of the NP-completeness proof is a
relatively straightforward reduction of the NP-complete 
set covering (\Sc ) problem. \Sc\ is defined as follows
\vspace{-.1in}
\begin{quote}
``...the set-covering problem consists of a
finite set $\cal F$ and a family $\cal N$ of subsets of $\cal F$, such
that every element of $\cal F$ belongs to at least one subset of $\cal
N$...We say that a subset $\cal S$ $\in$ $\cal N$ covers its elements.
The problem is to find a minimum-size subset ${\cal P} \subseteq {\cal
  N}$ whose members cover all of $\cal F$.'' (\cite{cormen90}, p.974,
we have substituted variable names analogous to those in \cl ).  
\end{quote}
\vspace{-.1in}
\cl\ can be viewed as extending \Sc\ in two ways.  First, it
substitutes features for integers.  Features introduce the possibility
of clashes between elements.  This makes the process of choosing
$\cal{P}$ more complex since one must take into account features
clashed with as well as features covered.  Second, intuitively, \cl\ 
loosens the restriction that all non-{\tt ?}-elements in $\cal{F}$
must be covered.  However, our formalization technically requires
complete coverage but retains the ability to represent incomplete
covers since listing a singleton set in the object class is equivalent
to listing an uncovered feature.

A polynomial time approximation algorithm exists for \Sc\ and the
solutions it produces are guaranteed to be close to the optimal
solution.  More formally, the ratio of the size of the approximate
solution produced by the algorithm to the optimal solution is bounded
by the natural logarithm of the size of the set being covered
\cite{lovasz75}: $|{\cal P}_{approx}|/|{\cal P}_{optimal}| \leq\ 
ln|{\cal F}| + 1$.  This approximation algorithm is greedy: at any
given point, it picks the subset that can cover the most features at
that time and it never goes back on this choice.  Greedy algorithms
tend to be fast but, since they cannot backtrack, they can make local
choices that prevent globally optimal solutions.  Because of the
similarities between \Sc\ and \cl , it seems likely that a greedy
approximation algorithm will produce good solutions for \cl .  Just
such an algorithm, \gcl , is listed below.  \vspace{-.1in} {\tt
  \footnotesize{
\begin{tabbing}
aaaa\=aaaa\= aa\= aa\= aaaa\= aaaa\= aaaa\= aaaa\= \kill
1.\> ${\cal F}_{temp}$ := $\{ [a,v] | [a,v] \in {\cal F} 
                                                  \wedge v \neq$ ?$\}$\\
2.\> $\cal P$ := $\emptyset$\\
3.\> ${\cal F}_{clash}$ := $\emptyset$ \\
4.\> while ${\cal F}_{temp} \neq \emptyset$\\
5.\> \> select ${\cal S} \in {\cal N}$ that maximizes 
        $|{\cal S} \cap {\cal F}_{temp}| - |clash({\cal F},{\cal S}) -
                                                   {\cal F}_{clash}|$\\
6.\> \> ${\cal F}_{temp}$ := ${\cal F}_{temp} - ({\cal S} \cup
                                            clash({\cal F},{\cal S}))$ \\
7.\> \> ${\cal F}_{clash}$ := ${\cal F}_{clash} \cup 
                                    clash({\cal F},{\cal S})$\\
8.\> \> $\cal P$ := ${\cal P} \cup \{{\cal S}\}$\\
9.\> return \\
10.\> list $\cal P$ and ${\cal F}_{clash}$ 
     in the data structure for the object\\
\end{tabbing}
}}
\vspace{-.2in}
\gcl\ takes as input the set of features of the object being inserted,
$\cal F$, and a set of sets of features, $\cal N$, that represents the
classes of the hierarchies.  It produces as output, a list of
superclasses and a list of features that must be listed locally for
the object.  During each iteration of its main loop (lines~4-9),
it picks the most suitable superclass (line~5).  The features
from $\cal F$ that can be inherited from the new superclass $\cal S$
combined with those that must be listed to block incorrect inheritance
are subtracted from $\cal F$.  This loop is repeated until there is no
class in $\cal N$ from which more features can be inherited than must
be listed to block incorrect inheritance.  The algorithm returns the
superclasses along with features that must be listed to block
incorrect inheritance.  To pick the most suitable superclass, a
subroutine is called for each class remaining in $\cal N$ that
computes the difference between the number of features remaining in
$\cal F$ that can be inherited from the class and the number of new
features that would be incorrectly inherited.\footnote{For some
  applications, features may not be equally important with respect to 
  insertion.  In such situations, features could be assigned
  different weights to account for the difference and in line~5 such
  weights would produce different payoffs and penalties.}
%\vspace{-.2in}
\section{\label{analysis}Analysis of the Algorithm}
%\vspace{-.1in}
As expected, \gcl\ runs in time polynomial in the size of the encoding
of the input pair \la $\cal F$,$\cal N$\ra , more specifically,
$O(min\{|{\cal F}_{non-?}|,|{\cal N}|\}|{\cal F}_{non-?}||{\cal N}|)$
where ${\cal F}_{non-?}$ represents the set of non-{\tt ?}-features in
$\cal F$.  The space complexity is also polynomial in the size of the
input: $O(max\{|{\cal F}_{non-?}|,|{\cal N}||{\cal S}_{max}|\})$ where ${\cal
  S}_{max}$ is the largest element of $\cal N$.

\gcl\ has been implemented and tested on a number of DATR
\cite{evens90} lexical hierarchies.  Testing consisted of enumerating
the features of a small number of words by hand, inserting the words
into the hierarchies by hand, running \gcl\ on the words, and then
comparing the insertions.  \gcl\ was extremely fast and produced
insertions that in most cases were very close to those done by hand.
Although these results are encouraging, we can, at this point, offer
neither extensive experimental results showing good performance nor a
theoretical proof of a logarithmic ratio bound for \gcl .  Instead, we
will discuss the basic situation that causes sub-optimal results to be
produced by greedy algorithms.  In addition, we will discuss how the
algorithm performs with respect to two standard problems for default
inheritance hierarchies taken from \cite{touretzky86}: redundant links
and nixon diamonds.  A system that utilizes default inheritance
hierarchies must address these problems since, if left unattended,
they can produce inconsistencies and/or unexpected behavior in the
system.

An example of the basic sub-optimal solution producing situation is
presented below.
%\enumsentence{\label{basic}\footnotesize{
\begin{myexample}\label{basic}
\begin{tabbing}
aaa\= \kill
\> A: [a1,v1][a2,v2][a3,v3][a4,v4] \\\null
\> B: [a1,v5][a2,v2][a3,v20][a7,v7][a9,v12] \\\null
\> C: [a3,v3][a4,v4][a6,v6] \\\null
\> $\cal F$ = \{[a1,v1][a2,v2][a3,v3][a4,v4][a5,v5][a6,v6]\} 
\end{tabbing} 
\end{myexample}
The problem here is that class A is seductive; it covers a large
number of features from $\cal F$, more than B or C.  However, if A is
chosen it is still necessary to pick B and C or the two appropriate
singleton sets to cover the features not covered by A: [a5,v5] and
[a6,v6].  If an algorithm can withstand the temptation of class A and
instead go with B or C, this algorithm is rewarded by simply choosing
the other class (either B or C) to cover the rest of $\cal F$. However,
\gcl\ falls for sets like A every time.  One can tweak the selection
criteria for the winning node (line~5) to produce an optimal
solution for any given instance.  However, in general, it is always
possible to come up with an instance for which these new criteria
produce a sub-optimal result.  This is due to the fact that the
central characteristic of the algorithm stays the same: decisions are
made with only local information and no backtracking is performed.
Note that such problematic instances do not seem to cause \gcl\ to
produce wildly incorrect solutions, simply slightly sub-optimal ones.

\begin{figure}[htbp]
% \centerline{\psfig{figure=/u/light/inserter/aaai/redun1.ps,height=1.2in}}
  \centerline{\psfig{figure=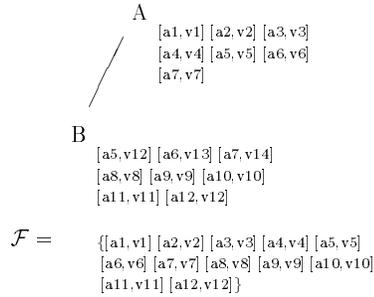,height=1.6in}}%1.2
  \caption{Redundant link producing situation}
  \label{redun1}
\end{figure}

Another situation that is problematic for \gcl\ is illustrated in
Fig.\ref{redun1}.
The compiled-out set, $\cal N$, (minus the singleton sets) for this
hierarchy is listed in (\ref{redunN}) where the first set corresponds
to A and the second to B.  
%\pagebreak
%\enumsentence{\label{redunN} \footnotesize{
\begin{myexample}\label{redunN}
\begin{tabbing}
aaa\=aaaa\=a\=a\=a\kill
\>$\cal N$ =\> \{\{[a1,v1] [a2,v2] [a3,v3] [a4,v4] [a5,v5] [a6,v6] [a7,v7]\}\\\null
\>\>\>           \{[a1,v1] [a2,v2] [a3,v3] [a4,v4] [a5,v12] [a6,v13] [a7,v14]\\\null
\>\>\>\>           [a8,v8] [a9,v9] [a10,v10] [a11,v11] [a12,v12]\}\}
\end{tabbing}
\end{myexample}
During the first iteration of the main loop of \gcl , A will be
selected to be a superclass since it will have a score of 7 whereas B
will only have a score of 6.  In the next round, however, B will be
chosen since it will then have a score of 2 which is higher than the
scores obtained by the singleton sets.  At this point all the features
of $\cal F$ will be covered.  Thus, the insertion for the object will
list A and B as parents and the following features [a5,v5], [a6,v6],
[a7,v7].  The link from the object class to A is known as a redundant
link because there is already a path to A through B.  However, notice
that the features that end up on the object class are the correct
ones; no inconsistencies exist.  This is a result of the way clashes
are handled.  Notice also that the solution is sub-optimal.  The
optimal insertion would list the same three features but only inherit
from B; the link to A could be removed without changing the features
of the object class.  Contrary to our earlier claim that each class in
a default hierarchy should be viewed as the set of features that can
be inherited from this class, it appears that Fig.\ref{redun1}
illustrates a case where the inheritance links between classes are
relevant for insertion.  One approach to this problem is to use
weights on features in the sets of $\cal N$ to encode the relevant
hierarchical structure of the classes.  These weights would be added
during the compilation process; the weight of a feature
would start at 1 in the class where it originates and would then
increase slightly each time it is passed down to a lower class.  
In line~5 of the algorithm, the weights would cause
 lower classes would be favored over
higher ones.  Alternatively, a post-processing algorithm could be
used to search for and eliminate redundant links.  In any case,
redundant links are a problem for \gcl\ as it currently stands.

The second standard problem for default inheritance hierarchies occurs
when an object class inherits from two classes each of which has a
different value for the same attribute.  This problem is known as the
nixon diamond and the standard example is that Richard Nixon was both a
Republican and a Quaker.  Republicans believe military force has to be
used sometimes whereas Quakers do not.  Thus, it is
unclear what features the object class Nixon should have.  The
question here is whether \gcl\ ever produces such situations.  The
answer is no.  Because clashes are handled by listing locally the
correct feature and because $\cal F$ is filled out by {\tt
  ?}-features, it is not possible for a Nixon diamond to result from a
\gcl\ insertion: when inserting Nixon, regardless of which class is
chosen first, [miluse,?] will be listed in the Nixon object class,
thereby blocking any inheritance of a feature with the
miluse attribute.
%\vspace{-.2in}
\section{Conclusion}
%\vspace{-.1in} 
We started this paper by discussing informally the problem of deciding
where an object belongs in a feature-based default inheritance
hierarchy.  This is an important problem but one that has, to our
knowledge, received little attention in the literature.
We formalized this problem as \cl .  A crucial
aspect of this formalization is that the hierarchy is viewed as a set
of unrelated sets of features.  Two facts combine to support this
claim: i) a good insertion minimizes the space needed to store the
object and thus, a class is seen as an opportunity to replace the
listing of a number of features by a single superclass, ii) the
structure of a default hierarchy cannot be used to reduce the search
space of potential superclasses.  Because \cl\ is NP-complete, we
designed an approximation algorithm for it.  We showed that this
algorithm is efficient and that it performs well with respect to a
pair of standard problems for default inheritance. 

\section{Acknowledgements}

The ideas about insertion discussed here grew out of the need for a
way to move data from the ELWIS database to inheritance hierarchies.
The INSYST project prototype mentioned above is a first attempt at
implementing such a system.  The INSYST project was a joint effort
between Sabine Reinhard, Marie Boyle-Hinrichs, and myself.  Marie
Boyle-Hinrichs was responsible for the implementation of the
INSYST prototype.  I would like to thank both of them for their
efforts and their support of my own.

Both ELWIS and INSYST were developed at the University of T\"{u}bingen
under the supervision of Erhard Hinrichs.  I would like to thank him
for the opportunity to spend a summer in T\"{u}bingen working on these
projects.  I would like to thank Lenhart Schubert for reading many
versions of this report and providing invaluable comments.  I would
also like to thank Chris Barker, George Ferguson, Dafydd Gibbon,
Sabine Reinhard, and Mark Young for their comments on earlier versions
of this report.  Finally, I would like to thank Yenjo Han, Leonidas
Kontothanassis, Jeff Schneider, and Paul Dietz for their help with the
initial design and analysis of \gcl .

\bibliographystyle{plain}
%\bibliography{light}

\end{document}